\documentclass[letterpaper, 10 pt, conference]{ieeeconf}  % Comment this line out if you need a4paper

\usepackage{graphicx}
\usepackage{multicol}
\usepackage{subcaption}
\usepackage{threeparttablex}
\usepackage{amsmath, amsfonts} % assumes amsmath package installed
\usepackage{amssymb}  % assumes amsmath package installed
\usepackage{mathtools}

\IEEEoverridecommandlockouts                              % This command is only needed if 
\title{\LARGE \bf
Physically Consistent Humanoid Loco-Manipulation using Latent Diffusion Models
}

\author{Ilyass Taouil$^{1*}$, Haizhou Zhao$^{2*}$, Angela Dai$^{1}$, Majid Khadiv$^{1}$% <-this % stops a space
\thanks{*Equal Contribution}% <-this % stops a space
\thanks{$^{1}$Technical University of Munich, Munich, Germany. E-mail: name.lastname@tum.de}%
\thanks{$^{2}$Haizhou Zhao, Tandon School of Engineering, New York University (NYU), USA. E-mail: name.lastname@nyu.edu}%
}

\begin{document}

\maketitle

\thispagestyle{empty}
\pagestyle{empty}

%%%%%%%%%%%%%%%%%%%%%%%%%%%%%%%%%%%%%%%%%%%%%%%%%%%%%%%%%%%%%%%%%%%%%%%%%%%%%%%%
\begin{abstract}
This paper uses the capabilities of latent diffusion models (LDMs) to generate realistic RGB human-object interaction scenes to guide humanoid loco-manipulation planning. To do so, we extract from the generated images both the contact locations and robot configurations that are then used inside a whole-body trajectory optimization (TO) formulation to generate physically consistent trajectories for humanoids. We validate our full pipeline in simulation for different long-horizon loco-manipulation scenarios and perform an extensive analysis of the proposed contact and robot configuration extraction pipeline. Our results show that using the information extracted from LDMs, we can generate physically consistent trajectories that require long-horizon reasoning.
\end{abstract}

%%%%%%%%%%%%%%%%%%%%%%%%%%%%%%%%%%%%%%%%%%%%%%%%%%%%%%%%%%%%%%%%%%%%%%%%%%%%%%%%
\section{INTRODUCTION}
\label{sec:introduction}

It has been long argued that humanoids are the best platform to replace humans in repetitive and dangerous tasks, because of the similarities in their morphologies. However, the complexity of these platforms poses significant challenges that have hindered the progress and we still do not see humanoid robots reliably doing real-world tasks. In particular, humanoids are high-dimensional systems with highly unstable dynamics (compared to wheeled and four-legged robots) which renders their planning problem highly challenging. Furthermore, performing any reasonable loco-manipulation task requires a long-horizon reasoning procedure and none of the existing methods can scale to such problems. The similarity between the human and humanoid morphologies can come to rescue in such a case, as the robot can imitate the behavior of humans doing the same task. Thanks to the recent advances in generative models, it is nowadays possible to generate a desired human behavior from text prompts. While the outputs of these models do not respect the geometrical and physical constraints of the real world, they can guide the existing optimization frameworks to find physically consistent motions quickly. 

In this paper, we develop a framework to rapidly synthesize plausible 3D human-object interaction scenes using latent diffusion models (LDMs), without the need for ad hoc heuristics or 3D richly annotated data, and use the retargeted motion inside a whole-body trajectory optimization (TO) formulation to generate physically and geometrically feasible motions.
The main contributions of this work are as follows:

\begin{itemize}
    \item We introduce, to the best of our knowledge, the first pipeline that plans both contacts and robot configurations for humanoid loco-manipulation using LDMs.

    \item We integrate our proposed robot configuration and contact planner within a whole-body TO formulation to generate physically consistent trajectories.

    \item We validate our approach in simulation on two challenging long-horizon scenarios, and perform an extensive analysis with various baselines. 

\end{itemize}

\begin{figure}
     \vspace{6pt}
     \centering
     \begin{subfigure}[b]{0.239\textwidth}
         \centering
         \includegraphics[width=\textwidth]{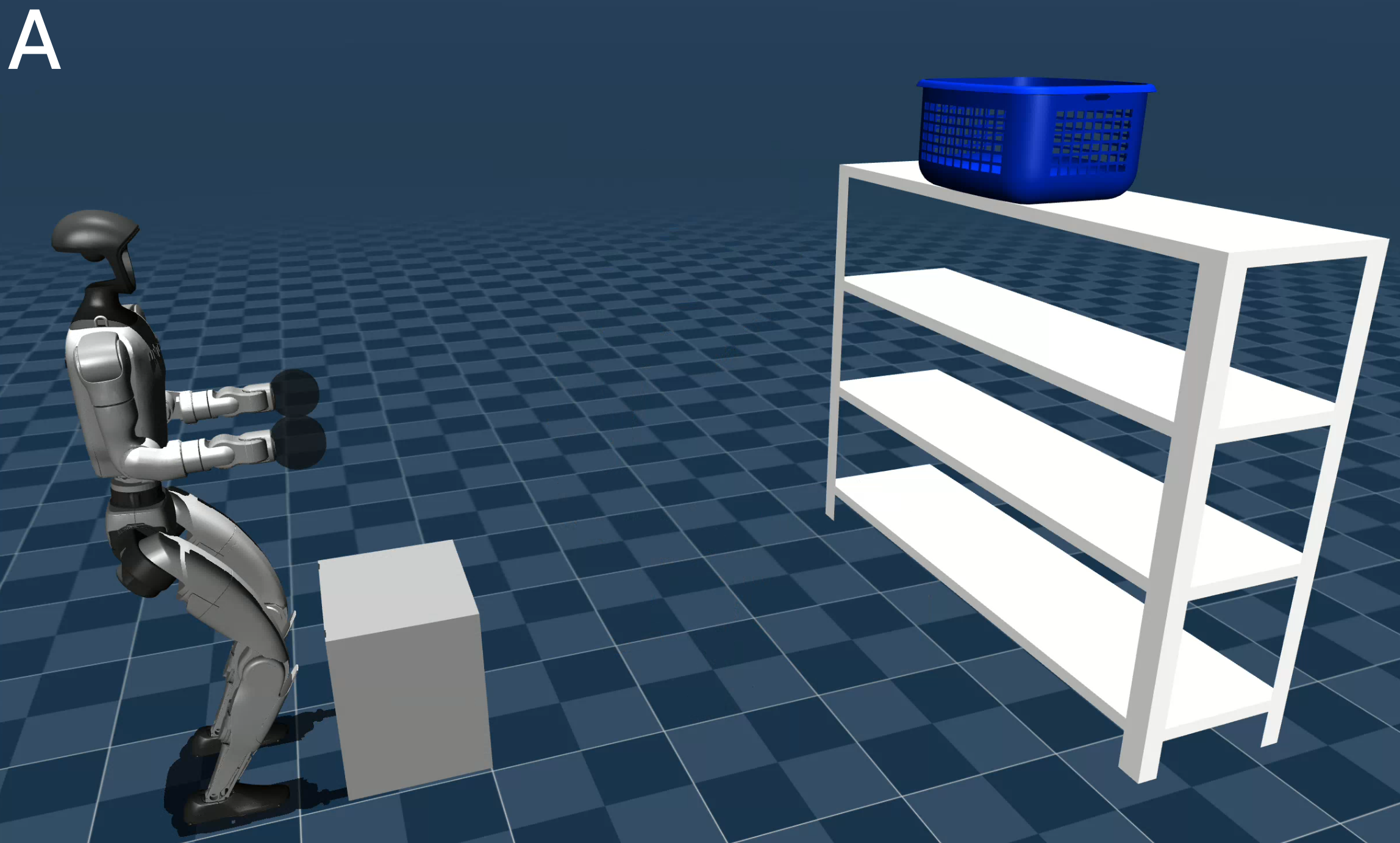}
     \end{subfigure}
     \begin{subfigure}[b]{0.239\textwidth}
         \centering
         \includegraphics[width=\textwidth]{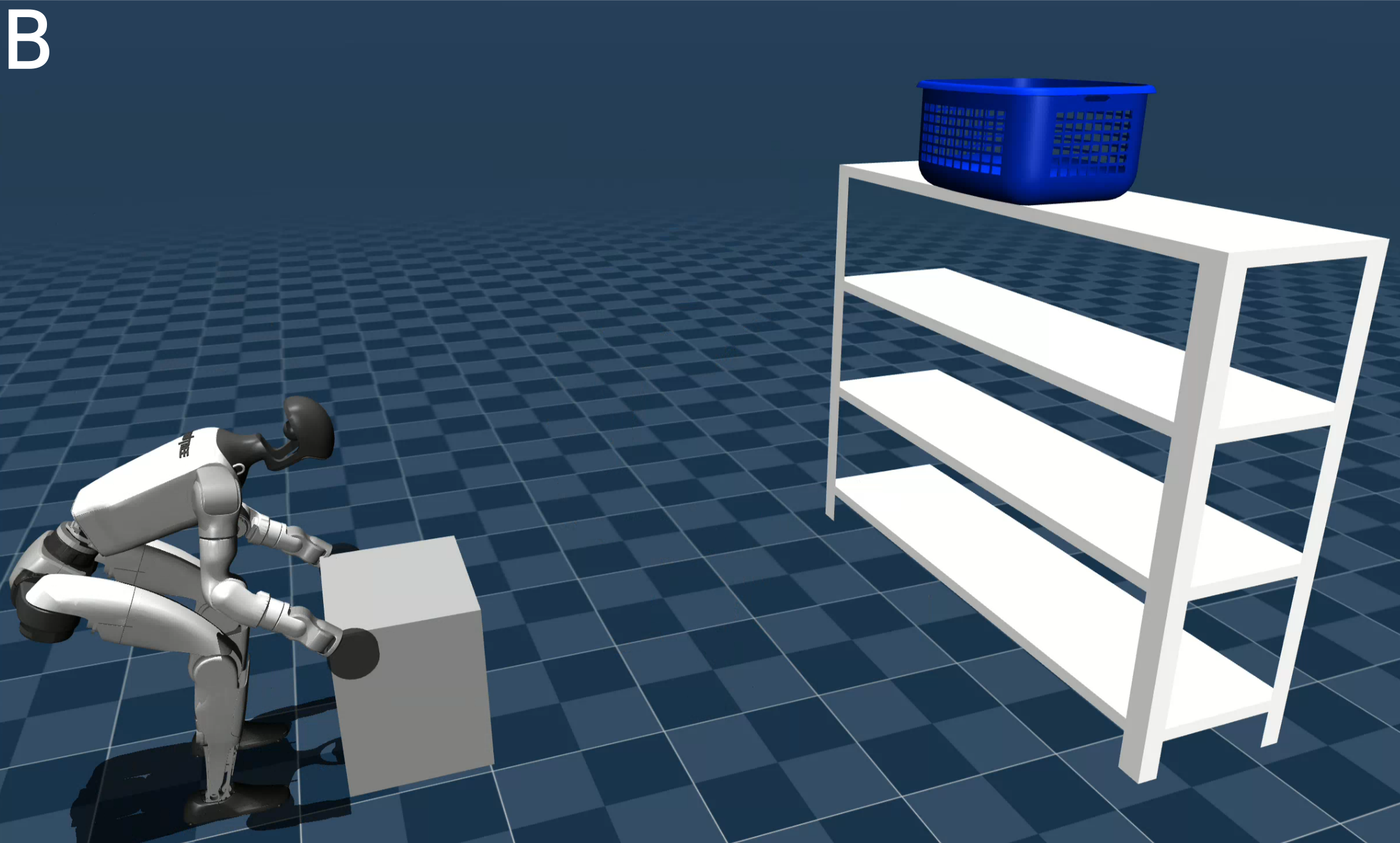}
     \end{subfigure}

     \vspace{3pt}
     
     \begin{subfigure}[b]{0.239\textwidth}
         \centering
         \includegraphics[width=\textwidth]{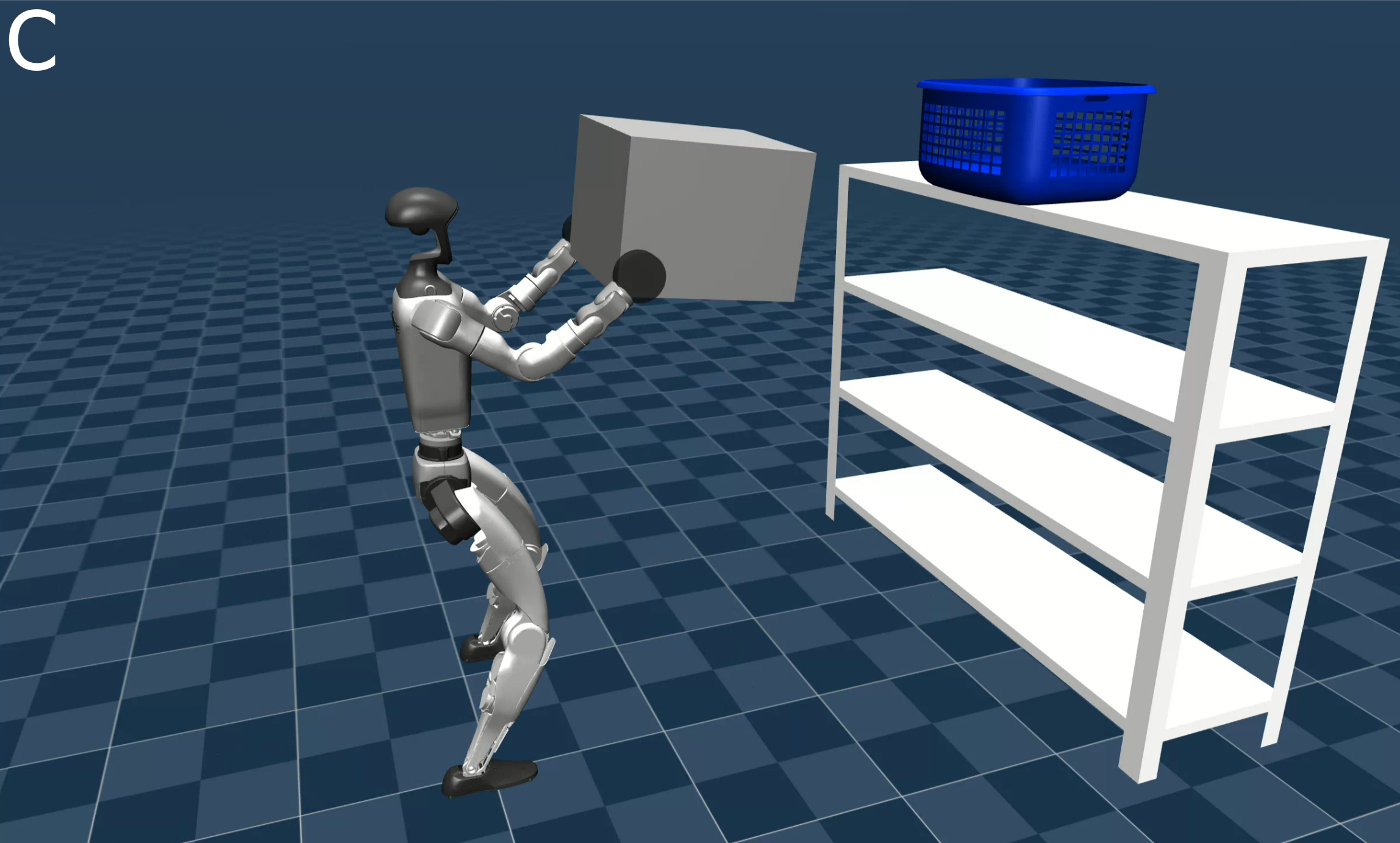}
     \end{subfigure}
     \begin{subfigure}[b]{0.239\textwidth}
         \centering
         \includegraphics[width=\textwidth]{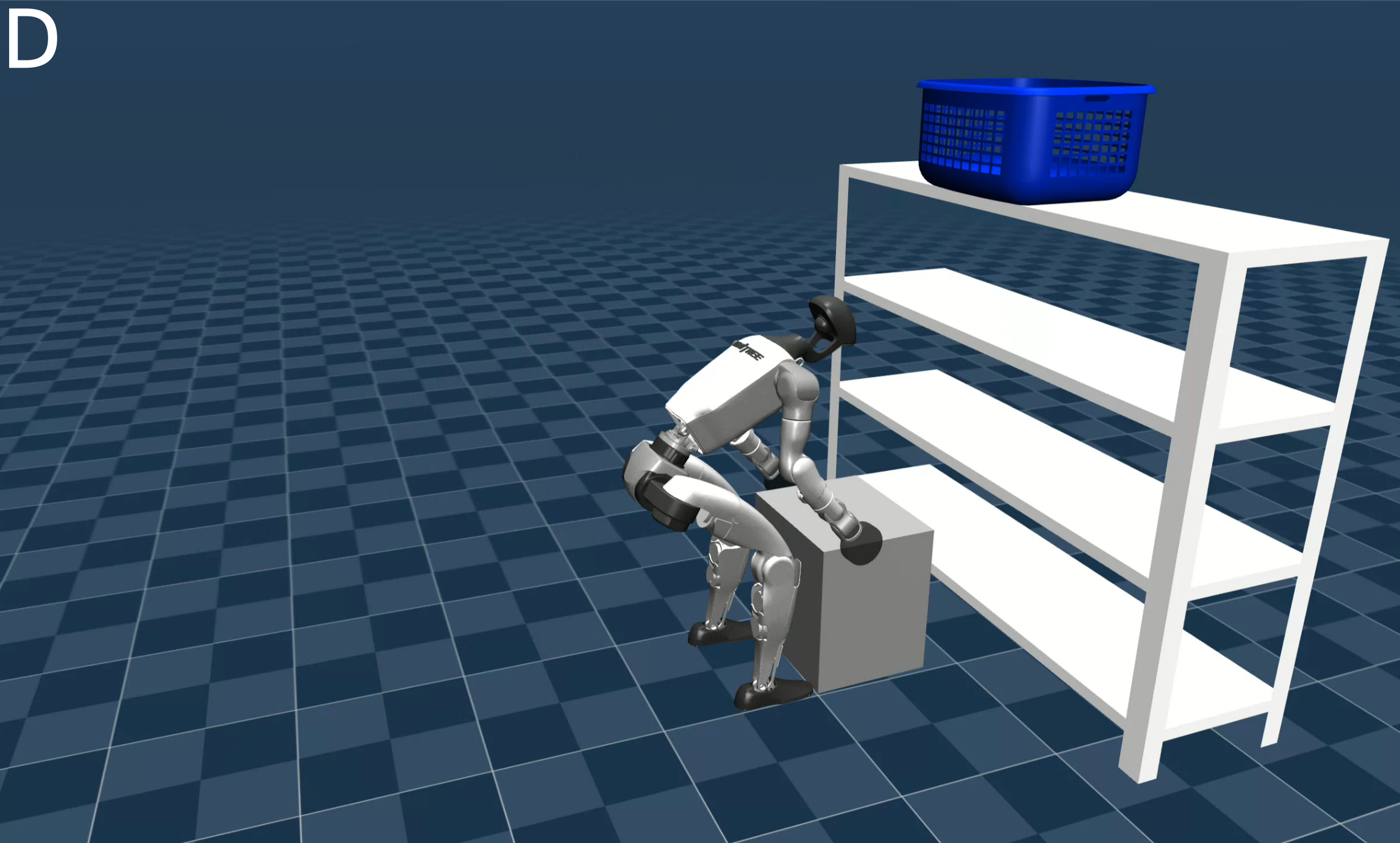}
     \end{subfigure}

     \vspace{3pt}
     
     \begin{subfigure}[b]{0.239\textwidth}
         \centering
         \includegraphics[width=\textwidth]{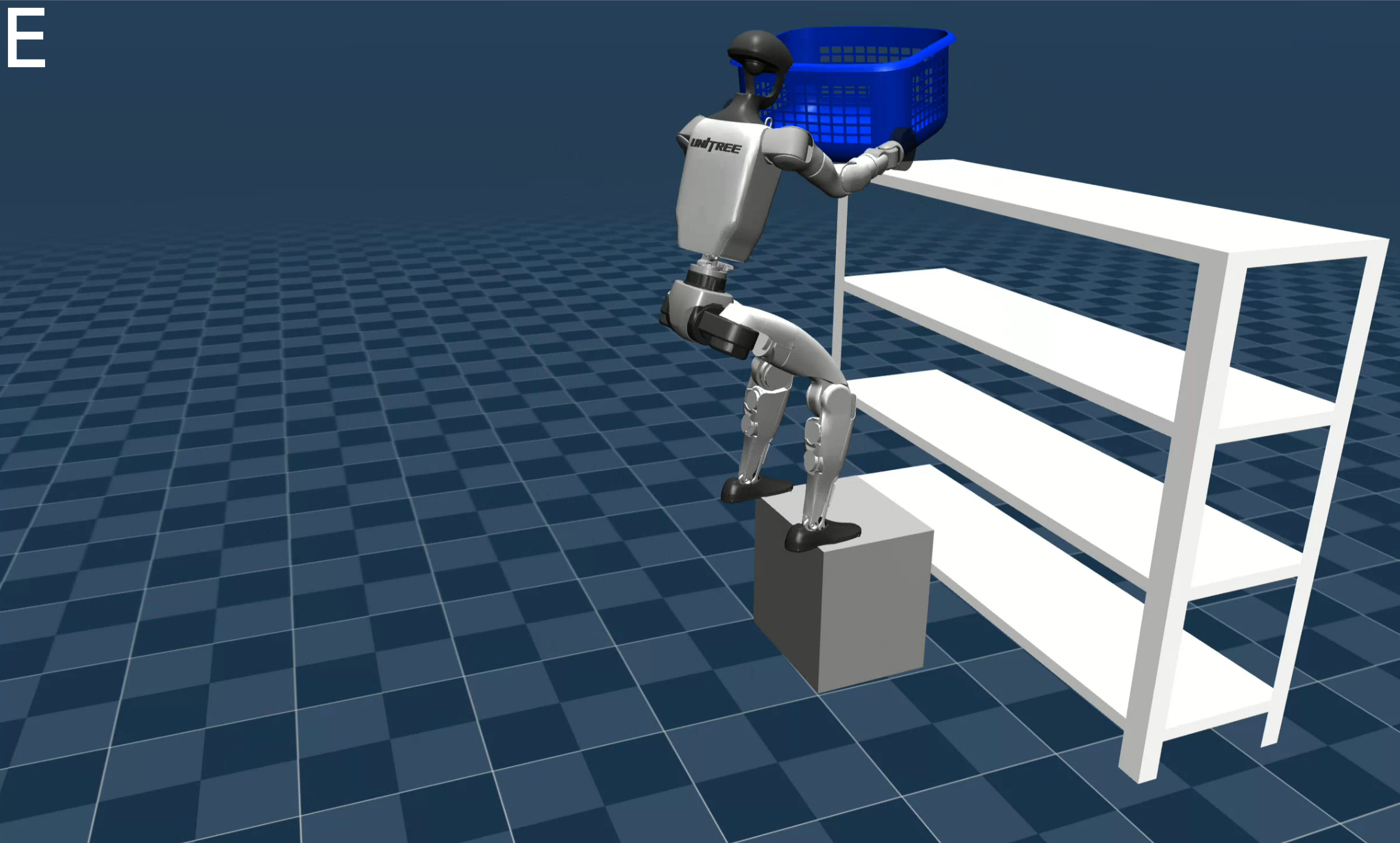}
     \end{subfigure}
     \begin{subfigure}[b]{0.239\textwidth}
         \centering
         \includegraphics[width=\textwidth]{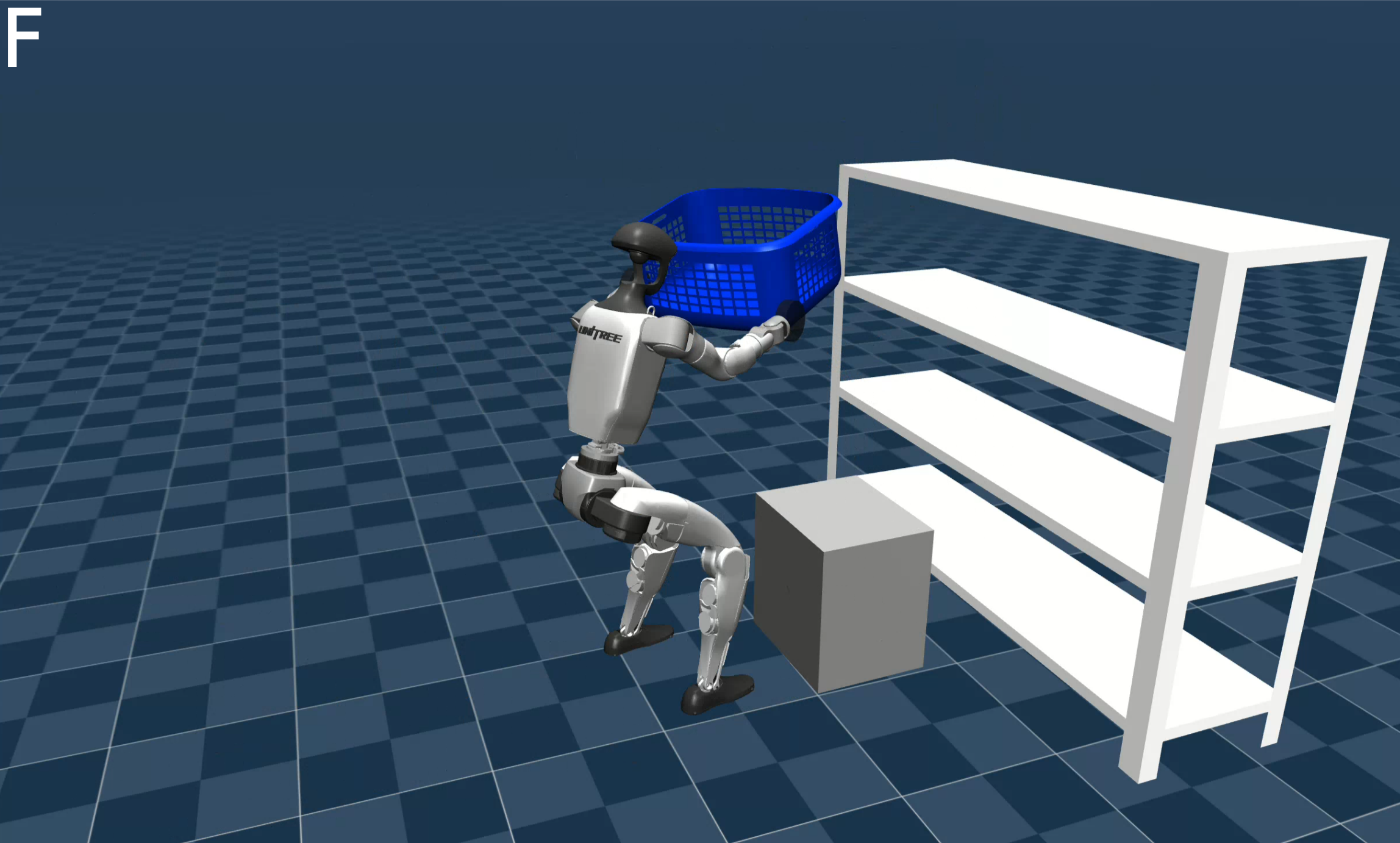}
     \end{subfigure}
    \vspace{-5mm}
    \caption{A loco-manipulation task achieved with our approach.}
    \vspace{-7mm}
    \label{fig:example}
\end{figure}%

\section{Related work}
\label{sec:related_works}

%\subsection{Planning \& Control for Loco-Manipulation}
Classical approaches for planning and control of loco-manipulation for humanoids consider the effect of manipulated objects on the locomotion system as a disturbance~\cite{bouyarmane2012humanoid,penco2019multimode,thibault2022standardized,li2023multi}. However, for general loco-manipulation problems, concurrent consideration of both locomotion and manipulation is essential. To reduce the complexity of the holisitc loco-manipulation planning, more advanced approaches relied on splitting the system into simpler coupled dynamical systems~\cite{settimi2016motion}, using heuristics to separate zones in which locomotion or loco-manipulation or manipulation occurs~\cite{ferrari2017humanoid}, splitting the object path planning and locomotion planning problems~\cite{murooka2021humanoid}, or using a predefined contact sequence~\cite{ferrolho2023roloma}.~\cite{sleiman2023versatile}
% demonstrated a complete pipeline for autonomously generating loco-manipulation plans. They 
used a hierarchy of optimal controllers to perform loco-manipulation automatically, augmenting the locomotion problem with logic predicates for manipulation~\cite{toussaint2018differentiable}. However, they demonstrated only quadrupedal loco-manipulation with single arm which is simpler than a humanoid with two arms. 

There have been recent efforts on the use of Deep Reinforcement Learning (DRL) for loco-manipulation tasks in the real world~\cite{fu2023deep,jeon2023learning,portela2024learning,ha2024learning}. However, these approaches are limited to very simple manipulation tasks with a quadruped and cannot reason about the complex long-horizon humanoid loco-manipulation tasks.  
% A complete hierarchical architecture based on sampling-based path planning, diffusion-based kinematic planning, and DRL-based tracking control showed interesting humanoid loco-manipulation animation behaviors~\cite{xie2023hierarchical}. However, not including the system dynamics in the planning stage can highly limit the feasibility of motions for real-world loco-manipulation systems. 
Recent advances in imitation learning have shown promise in generating loco-manipulation policies from teleoperation demonstrations~\cite{seo2023deep,ze2024generalizable}. However, generating teleoperated demonstrations for humanoid robots is extremely difficult compared to other manipulation settings~\cite{fu2024mobile}, as the system is highly unstable and can easily fall down.~\cite{liu2024opt2skill} used TO to generate demonstrations that are then imitated using DRL. However, TO is a local approach and would fail to generate long-horizon trajectories that require reasoning.

% \subsection{Leveraging Priors from Diffusion Models for Perception}

% Recent advances in powerful vision-language models~\cite{esser2024scaling, flux} have inspired efforts to transfer their generalized synthesis capabilities into a variety of other perception-based tasks.
% These include 2D panoptic segmentation~\cite{xu2023odise}, 3D semantic segmentation~\cite{rozenberszki2022language} and 3D scene generation~\cite{hoellein2023text2room}.
% More recently, GenZI~\cite{li2024genzi} pioneered an approach to distill high-level interaction priors from text-to-image based diffusion models to synthesize 3D human-scene interactions.~\cite{kim2025beyond} further employs 2D diffusion models to discover affordances in 3D human-object interactions.

\section{Method}
\label{sec:method}

\begin{figure*}[ht]
    \centering
    \includegraphics[width=.9\linewidth]{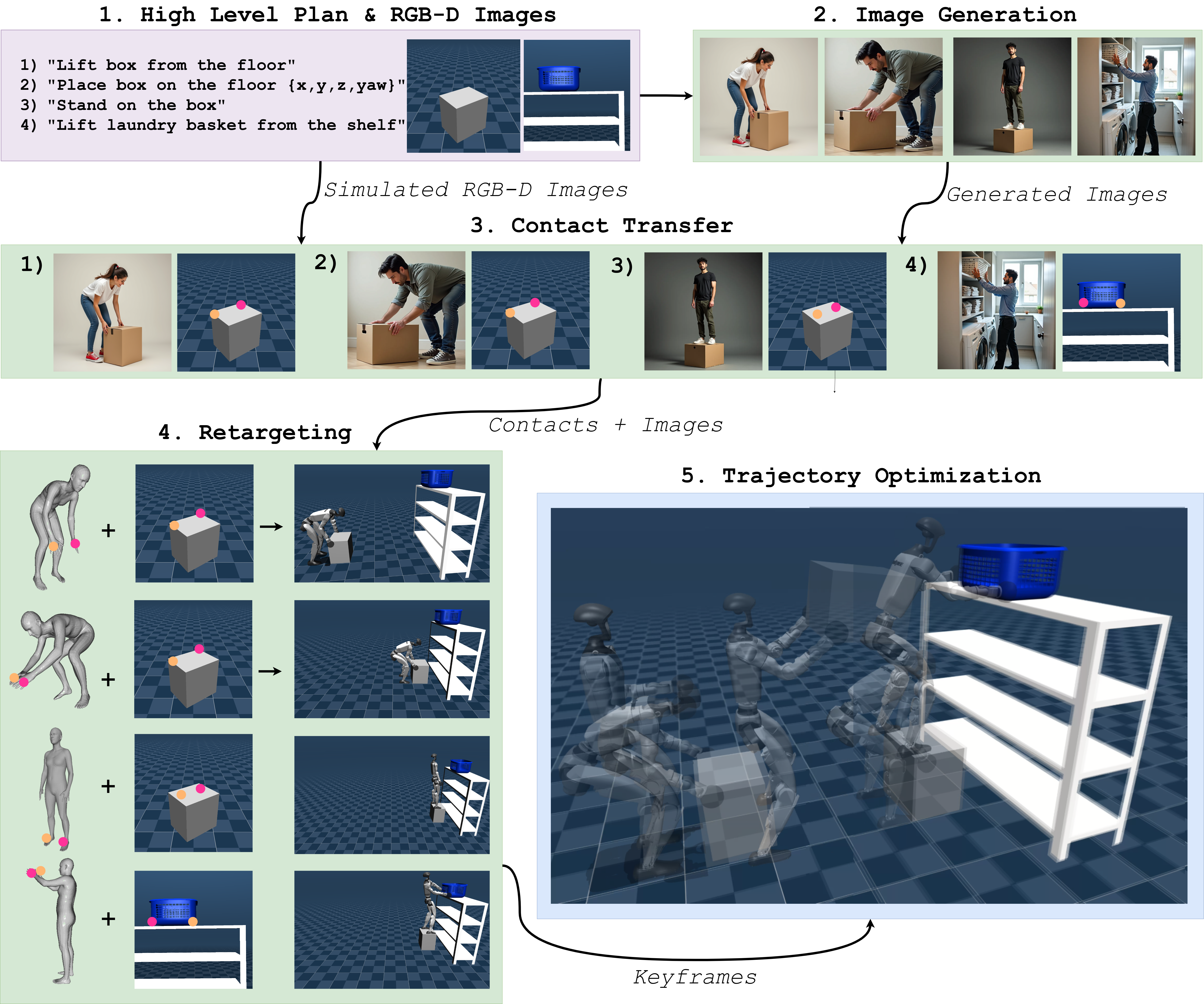}
    \caption{Pipeline overview.}
    \vspace{-7mm}
    \label{fig:pipeline}
\end{figure*}

In this section, we present our approach to plan contacts and robot configurations to guide a TO procedure for arbitrary long-horizon humanoid loco-manipulation tasks. Our approach does not rely on task-specific heuristics or 3D interaction datasets. Instead, we propose a pipeline that introduces an optimization-based approach that leverages LDMs to generate realistic human-object interaction 2D scenes, given a high-level description of the desired interactions. These 2D RGB scenes are used to extract the contact locations and robot configurations that are later used by TO (Section \ref{sec:trajectory_optimization}). The pipeline overview is illustrated in Fig.~\ref{fig:pipeline}.

\subsection{Planning Contacts \& Robot Configurations}
\label{sec:extracting_configurations_and_contacts}

The planner receives high-level instructions $P$ (that can come from a language model) and RGB-D images of the objects in the scene as input. The high-level plan consists of ordered sequences of text prompts describing how to break down the long-horizon task. For tasks involving placement, we assume to receive the target 3D location and yaw of the object. The RGB-D images consist of the RGB frames $R_s$ and respective depth images $D_s$ of the objects to be manipulated. The output of the planner is the sequence of 3D contact locations $L$ and associated robot configurations $C$. The planning process consists of three main steps which are detailed in the sections below.

\subsubsection{Image Generation}
\label{sec:image_generation}

Given $P$, we use a state-of-the-art latent diffusion model (LDM)~\cite{flux} to generate a collection of images $R_g$ demonstrating how to accomplish the long-horizon task. The instructions consist of an ordered sequence of short text prompts describing very minimally the expected interaction with the objects in the scene, as shown in Fig.~\ref{fig:pipeline}. Using directly these prompts leads to images that do not depict a full-body person, which is essential for our pipeline. In fact, to extract the contact locations of the hand and feet, and the respective robot configuration the vast majority of the human body has to be visible in the generated image. Therefore, we automatically append a static set of words to the task prompts to generate a full-body person. The additional words are mainly a general description of the person's hair color and clothing style, which forces the LDM to generate the correct interaction but also a full-body. Given an instruction, we modify it in the following way:

"\textit{A scene of person \{\textbf{predicate}\}+ing \{\textbf{subtask prompt without predicate}\}. The person has dark hair and is wearing casual clothes such a shirt, jeans, and sneakers}."

\vspace{3pt}

where \{\textit{\textbf{predicate}}\} is the verb describing the task's action and is always the first word in the task prompt, while \{\textbf{task prompt without predicate}\} is what remains of the prompt after removing the predicate and it mainly describes the object to which the action applies and its position in the environment. The generated images are then fed to the contact transfer (Sec.~\ref{sec:contact_transfer}) and retargeting (Sec.~\ref{sec:retargeting}) modules.

\subsubsection{Contact Transfer}
\label{sec:contact_transfer}

\begin{figure}
    \centering
    \vspace{2mm}
    \includegraphics[width=\linewidth]{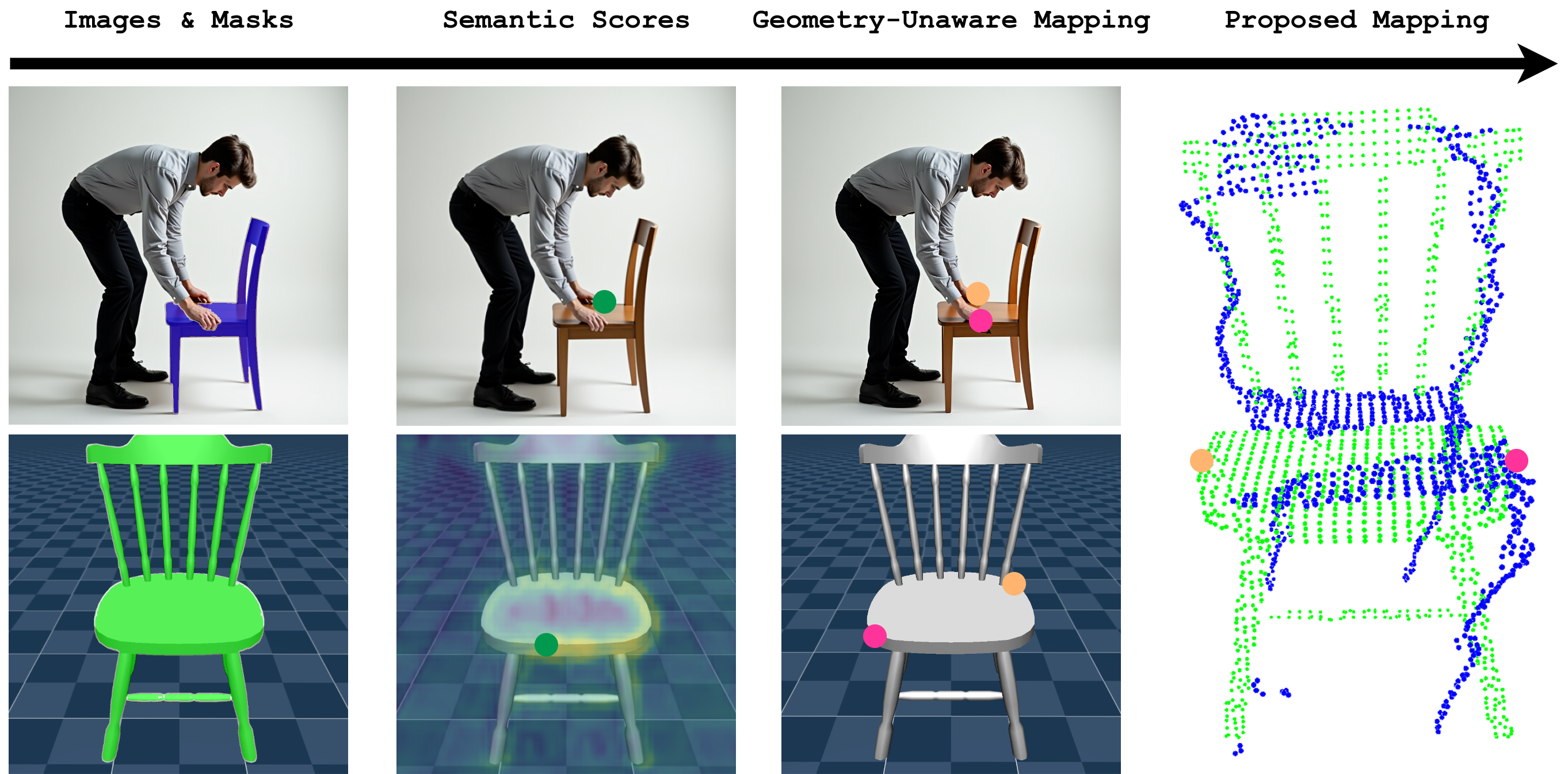}
    \caption{Contact extraction procedure.}
    \vspace{-7mm}
    \label{fig:contact_extraction_flow}
\end{figure}

The contact transfer stage extracts and maps the 2D contact information from the images in $R_g$ to the corresponding 3D contact locations in the scene to manipulate the object of interest for the task. To achieve this we use a three-step approach as shown in Fig.~\ref{fig:contact_extraction_flow}.

First, we compute the 2D semantic masks of the objects in $R_g$ and $R_s$. To do so, we use a Vision Language Model (VLM)~\cite{xiao2024florence} to perform open vocabulary object detection that returns the bounding box coordinates enclosing the objects. However, the obtained bounding box is not tight enough for the point cloud extraction later. Hence, we apply a visual segmentation foundation model~\cite{ravi2024sam} that further refines the VLM output and returns a per-pixel segmentation of the objects.

Second, using the objects' masks and the depth information we proceed to compute the objects' point clouds by performing a 2D to 3D lifting procedure. For the simulated images $R_s$ we have the ground truth depth $D_s$ from the simulated RGB-D camera in the MuJoCo~\cite{todorov2012mujoco} simulator and its correct camera intrinsic parameters. However, for $R_g$ we are missing both its depth estimate and correct camera intrinsics. This is because LDMs only output RGB images and do not adhere to a specific camera model during the image generation. Therefore, to estimate each generated image depth we leverage a zero-shot metric depth geometric foundation model~\cite{hu2024metric3d} to obtain the estimated metric depth $D_g$. The missing camera intrinsics are computed using an empirical trial and error approach where we found that using the LDMs' image resolution as the focal lengths and half the focal lengths for the principal point offsets leads to a reasonable point cloud geometry, without too much distortion. Finally, we also apply a noise removal process to the point clouds to remove outliers.

Third, we use a sampling-based optimization approach that combines the semantic scores from a semantic-aware foundation model and the object geometries to transfer the 2D contact locations from $R_g$ to the 3D world. The task of finding correct semantic correspondences across images is a challenging one, and especially so in our case. This is mainly due to the fact that during the image generation process, we have limited control over the generated object properties, such as viewpoint, shape, and texture, leading to significant intra-class variation between the generated and the simulated objects. To solve the semantic correspondence problem, we use a semantic-aware foundation model~\cite{zhang2024telling} to obtain semantic matches between the images. However, solely relying on the model is not reliable, as the intra-class variation can be large leading to incorrect mappings that deteriorate the output trajectory (Sec.~\ref{sec:contact_ablation}). Therefore, we propose a sampling-based algorithm that refines the correspondences from the semantic-aware model using the objects' point cloud geometries. The underlying idea is that correct semantic matches should result in a good geometrical overlap between the objects' point clouds. Hence, we generate a semantic score pool for some sampled 2D points on the objects' masks found in $R_g$, such that for each sampled point we obtain the top N most plausible correspondences in the respective objects' masks found in $R_s$ from the model. To find the set of semantic matches that best aligns the objects' geometries, we formulate a sampling-based algorithm that searches randomly within the semantic pool. More precisely, given the sampled semantic correspondences we compute the respective 3D correspondences and solve for the rigid-body transform with the Singular Value Decomposition (SVD) algorithm. The transform is then further refined with the Iterative Closest Point (ICP) algorithm. We repeat this process for 10 iterations and pick the rigid-body transform that obtained the highest overlapping score between the objects' point clouds. In practice, we found that within 3 iterations the best transform is already found. Finally, we compute $L$ by applying the transform to the 3D lifted hand and feet 2D locations obtained by running a human pose estimator~\cite{shin2024wham} on the images in $R_g$. To avoid penetrations between the real object and $L$, we apply a simple heuristic to $L$ to make sure these are projected to the closest object's surface.

\subsubsection{Retargeting}
\label{sec:retargeting}

\begin{figure}
    \centering
    \includegraphics[width=\linewidth]{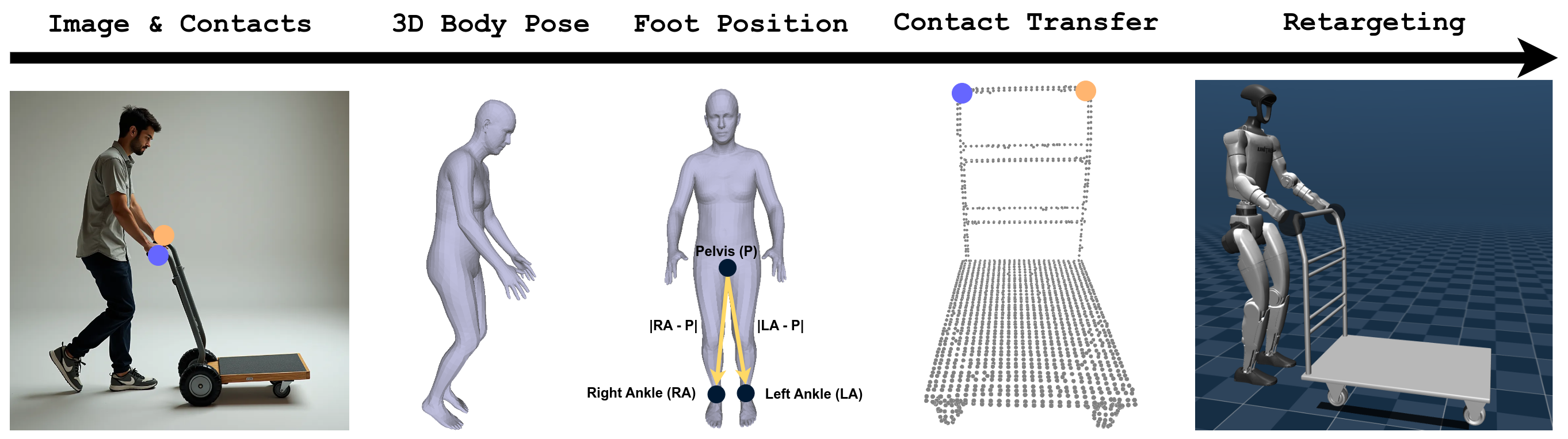}
    \vspace{-5mm}
    \caption{Keyframe extraction procedure.}
    \vspace{-7mm}
    \label{fig:retargeting}
\end{figure}

In the retargeting stage, we use the depicted humans in $R_g$ and $L$ to obtain $R$. $R$ consists of a 35D vector describing the robot's 6D base state and joint angles for each actuated joint of the system. Extracting this configuration is an important step in our proposed pipeline, as this complementary information to the contacts guides the TO to better local minima. However, we can't map directly the human configuration to the robot, due to differences in the number of degrees of freedom, limb length, and height. Therefore, we formulate an Inverse Kinematics (IK) based retargeting process that remaps the extracted human configurations from the generated images to a kinematically feasible robot configuration. Figure~\ref{fig:retargeting} shows an outline of the retargeting process.

First, we extract the human configuration we wish to remap, that consists of the human's joint angles, foot positions, and base orientation. To do so, we use WHAM~\cite{shin2024wham} a 3D human pose estimation module that returns the 3D joint positions and the joint orientations of a 2D image of a human body. For the joint orientations, as our humanoid has a subset of the degrees of freedom obtained from WHAM, we follow a similar approach to~\cite{fu2024humanplus}, where we only consider the human's joints that have a corresponding match on the robot. The foot positions are extracted from the 3D body model by computing the relative distance between the human's pelvis and the left and right ankles, however we only consider the planar coordinates as the foot height is set based on the task. The base orientation is obtained by first applying the rigid body transformation, computed during the contact transfer stage (Sec.~\ref{sec:contact_transfer}), to the 3D body model and then computing the relative orientation between the human's pelvis and the simulated object of which we know the full state. Finally, we apply the IK retargeting which uses as constraints the hand contact locations, feet contact locations, and feet pitch angle. The joint angles only act as a regularization term to guide the IK output towards a joint configuration similar to the human.

\section{Trajectory Optimization}
\label{sec:trajectory_optimization}

In this section, we outline our TO formulation and how the extracted contact locations and robot configurations are used within the formulation. 
We use the centroidal dynamics coupled with whole-body kinematics formulation similar to~\cite{dai_cenkindyn}.
% where the inputs are the joint velocities instead of torques or joint accelerations.
% The effectiveness of this choice has been proven via benchmark against the full dynamics in~\cite{justincompare}.
The semi-implicit Euler discretization of the dynamics is written as follows:
\begin{subequations}\label{eq:cenkindyn}
    \begin{align}
        \*{h}^+&=\*{h}^-+\begin{bmatrix}
            m\*{g}+\sum_{i\in\mathcal{C}}\*{f}_i\\
            \sum_{i\in\mathcal{C}}(\*{r}_i-\*c)\times\*{f}_i+\sum_{i\in\mathcal{C}}\boldsymbol{\tau}_i
        \end{bmatrix}\Delta t,\label{eq:cenkindyn_mom}\\
        \*{q}^+&=\*{q}^-+\*v^+\Delta t,\label{eq:cenkindyn_q}
    \end{align}
\end{subequations}

where $(\cdot)^{+/-}$ denotes the next or current value of $(\cdot)$, $\*h\in\mathbb{R}^6$ is the centroidal momentum, $\*g=[0,0,-9.81]^\top$ is the gravitational acceleration, $\mathcal{C}$ is the set of contacts, $\*r_i$ is the world frame contact point position, and $\*c$ is the center of mass (CoM). $\*f_i,\boldsymbol{\tau}_i\in\mathbb{R}^3$ are the forces and moments applied at the $i_{th}$ contact in the world frame, $\Delta t$ is the timestep, $\*q\in\mathbb{R}^{n_q}$ is the vector of generalized coordinates, and $\*v\in\mathbb{R}^{n_v}$ contains the velocities of the robot. For simplicity, Euler angles $\boldsymbol{\Theta}=[\phi, \theta, \psi]^\top$ are used for the floating-base rotation. The same dynamics in~\eqref{eq:cenkindyn} is also used for the objects.

\subsection{Contact Modalities}
\label{sec:contact_modalities}

% For loco-manipulation tasks, various contact modalities need to be considered, which we detail in the following subsection.

\subsubsection{Patch Contact}
\label{patch_contact}

Assuming the contact surface is perpendicular to the gravity (e.g., ground, table), the patch contact constraint for the $i_{th}$ contact is formulated as follows:
\begin{subequations}\label{eq:patch_contact_constr}
    \begin{align}
        (\*r_{i,z}-z_g)+\dot{\*r}_{i,z}/k_{z}+s_{i,z}=0,\label{eq:patch_contact_constr_z}\\
        \dot{\*r}_{i,xy}=\*0,\label{eq:patch_contact_constr_xy}\\
        [\*{I}_2,\*0_{2\times1}]\*R_{i,z}=\*0,\label{eq:patch_contact_norm}\\
        \*w_{i,xy}=\*0,\label{eq:patch_contact_constr_angular_vel}
    \end{align}
\end{subequations}

where~\eqref{eq:patch_contact_constr_z} constrains the distance between the contact point and the surface, $k_{z}$ is the constraint stabilization stiffness, and $s_{i,z}$ the slack variable.~\eqref{eq:patch_contact_constr_xy} enforces the non-sliding constraint.~\eqref{eq:patch_contact_norm} is the patch orientation constraint to ensure alignment with the surface, where $\*R_{i, z}$ is the z-basis of the contact patch rotation matrix $\*R_i\in SO(3)$ w.r.t. the reference surface.  Finally,~\eqref{eq:patch_contact_constr_angular_vel} makes sure the object does not start to rotate into the surface, where $\*w_i$ is the object's angular velocity. Note that the orientation constraint can be represented in other forms but~\eqref{eq:patch_contact_norm} has been the most robust one in practice.

The linearized friction cone as well as CoP constraintsin the local contact frame for square patches are simplified as:
\begin{subequations}\label{eq:patch_friction_cone}
    \begin{align}
        \mu{}_C\*f_{i,z}\ge|{}_C\*f_{i,x}|, \quad \mu{}_C\*f_{i,z}\ge|{}_C\*f_{i,y}|,\\
        {}_C\*f_{i,z}X\ge|{}_C\boldsymbol{\tau}_{i,y}|, \quad {}_C\*f_{i,z}Y\ge|{}_C\boldsymbol{\tau}_{i,x}|,\\
        \mu \: {}_C\*f_{i,z}\max\{X, Y\}\ge|{}_C\boldsymbol{\tau}_{i,z}|,\label{eq:fric_cone_ztorque}
    \end{align}
\end{subequations}

where $\mu$ is the friction coefficient, and $X$ and $Y$ are respectively the size of the patch in the x and y directions.

\subsubsection{Interactive Patch Contact}
\label{sec:interactive_patch_contact}

For moving contacts (e.g., robot-object), equations~\eqref{eq:patch_contact_constr_z} and~\eqref{eq:patch_contact_constr_xy} are changed to:
\begin{equation}\label{eq:interactive_patch_pos_constr}
    \*r_{i}-\*r_{j} + \*s_{ij}=\*0,
\end{equation}

where $i\in\mathcal{C}_A,j\in\mathcal{C}_B$ are the contact indices of the system $A, B$, and $\*s_{ij}\in\mathbb{R}^3$ is the stacked slack variables. The orientation constraint in~\eqref{eq:patch_contact_norm} and~\eqref{eq:patch_contact_constr_angular_vel} are changed to the relative orientation of $i$ w.r.t. $j$ as follows:
\begin{subequations}
    \begin{align}
        [\*I_2,\*0_{2\times1}](\*R_j^\top \*R_i)_z=\*0,\\
        \*w_i-\*w_j=\*0.
    \end{align}
\end{subequations}

We need to also enforce Newton's law of reaction:

\begin{equation}
    \*f_i+\*f_j=\*0,\quad \boldsymbol{\tau}_i+\boldsymbol{\tau}_j=\*0.
\end{equation}

In case the contact is prehensible, such as grasping the trolley's handles, we can simply omit the constraints in~\eqref{eq:patch_friction_cone} allowing arbitrary forces (assuming that the grasp remains firm).

\subsubsection{Nonholonomic Chassis Constraint}
\label{sec:nonholonomic_wheel_constraint}

In one of our experiments, the nonholonomic constraint of a trolley chassis is needed. All wheels are assumed to have pure rolling contacts such that the chassis can slide freely in the tangential direction. The front wheels are assumed omnidirectional and can rotate around the z-axis. Let $C_\text{ch}$ be the virtual chassis contact frame on the ground and located below the chassis platform center. The chassis contact constraint can be formulated as:
\begin{subequations}
    \begin{align}
    \text{for}~i_\text{ch}, C_\text{ch}:
   ~\eqref{eq:patch_contact_constr_z},\eqref{eq:patch_contact_norm},\eqref{eq:patch_contact_constr_angular_vel},\nonumber\\
    {}_{C_\text{ch}}\dot{\*r}_{w,y}=0\label{eq:chassis_y_vel},\\
    {}_{C_\text{ch}}\*f_{i_\text{ch}, x}=0\label{eq:chassis_x_force},
    \end{align}
\end{subequations}

where ${}_{C_\text{ch}}\dot{\*r}_{w,y}$ denotes the local lateral velocity of the rear wheels and $i_\text{ch}$ is the contact index of the chassis.~\eqref{eq:chassis_y_vel} disables lateral movement of rear wheels and~\eqref{eq:chassis_x_force} disables the tangential ground reaction force exerted on the chassis (in the direction of forward/backward movement).

\subsection{Task-generic Cost}

To avoid task-specific tuning, the stagewise cost $L_\text{stage}$ is designed to regularize the robot locomotion with minimum heuristics:
\begin{equation}
    L_\text{stage}:= b_\text{st} L_\text{st} + (1 - b_\text{st}) L_\text{wk} + L_\text{reg} + L_\text{slack},
\end{equation}

where $b_\text{st}$ is the Boolean flag indicating whether the robot is in a stance phase, $L_\text{st}, L_\text{wk}$ are respectively the cost of stance and walking phases, $L_\text{reg}$ is the common regularization term, and $L_\text{slack}$ is the penalty of the slack variables used for constraints such as~\eqref{eq:patch_contact_constr_z} and~\eqref{eq:interactive_patch_pos_constr}. All costs are in quadratic form $w||(\cdot)||_2^2$, where $w$ is the weight. 
The detailed description of all the terms as well as values used for the experiments are summarized in Table~\ref{tab:stagewise_cost}.

\begin{table}[t]
    \centering
    \caption{Stagewise Costs used for experiments}
    \begin{threeparttable}
    \renewcommand{\arraystretch}{1.1}
    \begin{tabular}{|l|l|l|}
         \hline
         \multicolumn{3}{|c|}{\textbf{Stance phase cost} $L_\text{st}$}\\
         \hline
         \textbf{Name} & \textbf{Equation} & \textbf{Weight} \\ 
         \hline
         CoM w.r.t. foot & $\*c_\text{all} - (\*r_{lf,xy}+\*r_{rf,xy})/2$ & 5e2 \\ 
         Leg symmetry & $[q_{p}^l-q_{p}^r; q_{r}^l; q_{r}^r; q_{y}^l+q_{y}^r;q_{k}^l-q_{k}^r]$ & 3e2 \\ 
         Foot yaw symmetry & $\*R_{lf,x}-\*R_{rf,x}$& 1e2 \\
         Base pose & $[\theta, \psi]^\top$ & 1e1 \\ \hline
         \multicolumn{3}{|c|}{\textbf{Walking phase cost} $L_\text{wk}$}\\
         \hline
         Foot clearance & $\*r_\text{sw, z} - \text{0.07}$ & 2e3 \\
         Leg roll & $[q_r^l,q_r^r]^\top$ & 1e1\\
         Base pose & $[\theta, \psi]^\top$ & 1e2 \\ \hline
         \multicolumn{3}{|c|}{\textbf{Common regularization} $L_\text{reg}$}\\
         \hline
         Force & $[\*f_i; \boldsymbol{\tau}_i],\forall i$ & 1e-6\\
         Base linear velocity & $\dot{\*r}_\text{base}$ & 1e-1 \\ 
         Base angular velocity & $\*w_\text{base}$ & 1 \\ 
         Leg velocities & $\dot{\*q}_\text{leg}$ & 5e-1 \\
         Arm velocities & $\dot{\*q}_\text{arm}$ & 1 \\
         Arm joint angles & $\*q_\text{arm}$ & 1 \\
         Base height & $\*r_{\text{base},z}-\text{0.7}$& 1e2 \\
         Timestep & $\Delta t - \text{2e-2}$ & 1e2 \\
         \hline
         \multicolumn{3}{|c|}{\textbf{Slack penalty} $L_\text{slack}$}\\
         \hline
         Slack in~\eqref{eq:patch_contact_constr_z},\eqref{eq:interactive_patch_pos_constr} & $s_{(\cdot)}$ & 1e5\\
         \hline
    \end{tabular}
    \begin{tablenotes}[para, flushleft]
        \textbf{Note}: $(\cdot)_{(l/r)f}$ is the left/right foot pose, $\*c_\text{all}$ is the CoM of the supported mass including the robot and the lifted objects, $\*r_{sw}$ is the swing foot position, $q_{(\cdot)}^{l/r}$ denotes the joint angle of the left/right hip pitch ($p$), hip roll ($r$), hip yaw ($y$), knee ($k$),
    \end{tablenotes}
    \end{threeparttable}
    \vspace{-5mm}
    \label{tab:stagewise_cost}
\end{table}

\subsection{Guiding Optimization with Keyframes}
\label{sec:guiding_optimization_with_keyframes}

\subsubsection{Cost Design with Keyframes}
\label{sec:cost_design_with_keyframes}

The full configuration of the robot and the contact locations generated from the planning module constitute a~\emph{keyframe}, which provides waypoints for a long-horizon loco-manipulation task. We add the following cost term for the keyframe robot pose:
\begin{equation}\label{eq:kf_base_costs}
    L_\text{kf}^\text{b}:=W_\text{kf}^\text{b}[(\*r_{\text{base}, z}-\*r_{\text{base}, z}^\text{kf})^2,||\boldsymbol{\Theta}-\boldsymbol{\Theta}^\text{kf}||^2_2]^\top,
\end{equation}
where $W_\text{kf}^\text{b}\in\mathbb{R}^{1\times2}$ is the cost weight for which we used $[100, 10]$ in this paper. $(\cdot)^\text{kf}$ denotes the corresponding value at a certain keyframe, $r_{\text{base}, z}$ is the $z$ component of the robot base, and $\boldsymbol{\Theta}$ is the aforementioned base orientation. The keyframes also indicate the desired relative foot position w.r.t. the object, which can then be used to compute a global position reference $\mathbf{r}_f^\text{des}$. It is added as an additional term to the corresponding stance phase cost $L_\text{st}$:

\begin{equation}\label{eq:kf_foot_costs}
L_\text{kf}^\text{f}:=W_\text{kf}^\text{f}||(\*r_{lf,xy}+\*r_{rf,xy})/2 - \mathbf{r}_f^\text{des}||^2_2,
\end{equation}

where $W_\text{kf}^\text{f}$ is the keyframe foot position weight for which we used 5e2 in this paper.

\subsubsection{Subtask-like Warm Start}
\label{sec:kf_warm_start} 

Our long-horizon loco-manipulation problem can be decomposed into several small-scale subproblems that solve subtasks separately. For instance, if a robot is commanded to move a box placed on the ground to a table, the subtasks should be: walking to the box, picking it up, walking to the table, and placing the box on it. For each subproblem, a keyframe is its initial state, which is conventionally regarded as a nominal state to initialize all shooting nodes. Similarly, when solving the original problem holistically, for the nodes corresponding to each subtask, we initialize their robot states to be the corresponding keyframes, if available. This can greatly accelerate and robustify the convergence of long-horizon optimization.

\subsection{Collision Avoidance}
\label{sec:collision_avoidance}

We introduce various techniques for different types of obstacles to generate more realistic motion:

\begin{itemize}
    \item Hard penetration constraint:
    \begin{equation}\label{eq:hard_penetration_constraint}
        p:=(d^2-1), p\ge0
    \end{equation}
    where $p$ is the scaled penetration, $d$ is the scaled distance w.r.t. the centroid of the penetrated body. When $d=1$, the two bodies in contact intersect at a single contact point.
    
    \item Non-smooth penalties of penetration:
    \begin{equation}\label{eq:nonsmooth_penetration_penalty}
        L_\text{pen}:=\bigg\{\begin{array}{ll}
            p^2, & \text{if}~p <0; \\
            0, & \text{otherwise}.
        \end{array}
    \end{equation}
    
    \item Homotopy-based penetration constraint and penalty:
    \begin{subequations}\label{eq:homotopy_collision}
        \begin{align}
            &\alpha \tilde{p}+(1-\alpha)p\ge0, \alpha\in[0,1],\label{eq:homotopy_constr}\\
            &L_\text{hom}:=w_\text{hom}\alpha^2, \label{eq:homotopy_penalty}
        \end{align}
    \end{subequations}
    where $\alpha$ is a homotopy parameter driven to zero by the penalty $L_\text{hom}$ in~\eqref{eq:homotopy_penalty} and $w_\text{hom}$ is the penalty weight for which we used 2e2 in this paper. $\tilde{p}$ is the second-order differentiable scaled penetration into an alternative body, while $p$ is that of the original body which is usually complex or non-smooth. It enables guiding the optimizer with a simpler smooth geometry, such as using an ellipsoid to guide solving for a cube obstacle. It has better numerical properties compared to ~\eqref{eq:hard_penetration_constraint}, while preserving relatively high accuracy. However,~\eqref{eq:homotopy_collision} is still more difficult to solve compared with~\eqref{eq:nonsmooth_penetration_penalty}.
\end{itemize}

In the experiments, we use combinations of the above to implement collision avoidance between the robot, the objects, and the environment.

\subsection{Overall Optimization Problem}
\label{sec:overall_optimization_problem}

The overall TO problem can be formulated as
\begin{subequations}
    \begin{align}
        \min~&\frac1N\sum_{i=0}^N [L_\text{stage}^i+L_\text{col}^i]+\sum_{j\in\mathcal{K}}L_\text{kf}^{\text{b},j}\\
        \text{s.t.}~&\quad\text{Dynamics}~\eqref{eq:cenkindyn}\label{eq:overall_dyn_constr}\\
        &\quad\text{Contacts}~\text{e.g.,}~\eqref{eq:patch_contact_constr},\eqref{eq:patch_friction_cone}\label{eq:overall_contact_constr}\\
        &\quad\text{Collision}~\text{e.g.,}~\eqref{eq:hard_penetration_constraint},\eqref{eq:homotopy_constr}\label{eq:overall_collision}
    \end{align}
\end{subequations}

where the decision variables are all $\*q,\*v,\*f,\boldsymbol{\tau}, \*s, \alpha, \Delta t$. $N$ is the number of shooting nodes, $\mathcal{K}$ is the set of keyframe indices, $L_\text{col}$ denotes the possible collision penalties including~\eqref{eq:nonsmooth_penetration_penalty} and~\eqref{eq:homotopy_penalty}. 

\section{Experiments}
\label{sec:experiments}

In this section, we present the results of applying our proposed pipeline for two different scenarios, each involving a long-horizon task. The first scenario (S1) consists of fetching a laundry basket placed on top of a shelf. As the basket is not easily reachable, the robot needs to move a box close to the shelf and step on top of it to be able to reach the basket. The second scenario (S2) consists of moving a box placed on top of a table using a trolley and then pushing the trolley. We used the MuJoCo simulation environment~\cite{todorov2012mujoco} and the Unitree G1 humanoid robot for all the visualization and comparisons.

First, we demonstrate the promise of our framework in generating physically plausible trajectories for both scenarios. We also compare our results to a TO baseline that is guided only through contacts from a semantic-aware foundation model. Note that without providing such information, it was impossible for TO to solve the tasks, as both tasks require long-horizon reasoning that is unfeasible to do with local TO. 

Second, we perform an ablation study on our proposed contact extraction process (Sec.~\ref{sec:contact_transfer}), where we compare a geometry-unaware contact transfer and our method to see how much our proposed contact refinement improves the resulting trajectories. 

Third, we carry out an ablation study on the effect of each component of the keyframe information (Sec.~\ref{sec:retargeting}) on the TO output. We refer the reader to the supplementary video for the additional qualitative results.

% \begin{figure}[t]
%     \centering
%     \includegraphics[width=\linewidth]{images/physically_plausible_qualitative.png}
%     \caption{Qualitative comparison between the TO output using our proposed pipeline and a naive approach for the laundry scenario (S1) and trolley scenario (S2).}
%     \vspace{-7mm}
%     \label{fig:punch_line}
% \end{figure}

\subsection{Physically Plausible Trajectories}
\label{sec:physically_plausible_trajectories}

We compare the output of the TO results (Sec.~\ref{sec:trajectory_optimization}) when using the output of our proposed planning pipeline (Sec.~\ref{sec:extracting_configurations_and_contacts}) and a naive approach. Our pipeline feeds both the refined contact information and the respective robot configuration extracted from LDMs to TO. The naive approach only gives the contact locations of the hands or feet, depending on the task, obtained directly from the semantic-aware foundation model to the TO. In both cases, we use a minimal set of collision penalties constraints~\eqref{eq:nonsmooth_penetration_penalty}. For S1, these are between 1) the robot knees/feet and the box when stepping onto and off it and 2) the basket and the shelf. For S2, we added the constraints between 1) the robot hands and the box using~\eqref{eq:nonsmooth_penetration_penalty}, 2) the robot hands and the trolley using~\eqref{eq:nonsmooth_penetration_penalty}, 3) the robot legs and the trolley using~\eqref{eq:homotopy_collision}, 4) between the knees when walking to the trolley using~\eqref{eq:hard_penetration_constraint}.

% Figure~\ref{fig:punch_line} shows the snapshots of some trajectory instances when using a naive approach compared to our proposed method. We can clearly see that TO with naive contacts from the foundation model incurs a lot more negative self-collisions and robot-object collisions compared to TO using our planning procedure. In fact, by providing reasonable contacts and robot configurations, we guide TO to find collision-free motions that also respect physics.

\begin{figure}
     \vspace{6pt}
     \centering
     \begin{subfigure}[b]{0.239\textwidth}
         \centering
         \includegraphics[width=\textwidth]{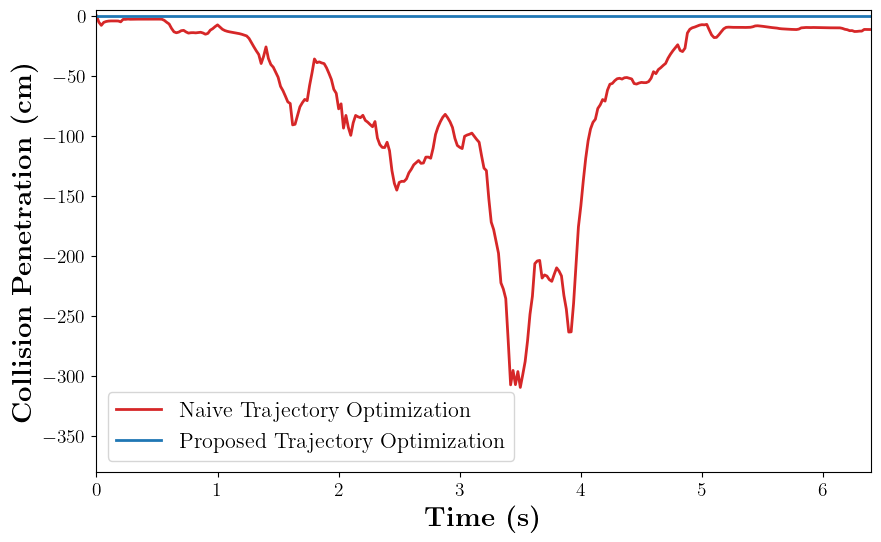}
         \caption{S1 w/ collision penalties}
     \end{subfigure}
     \begin{subfigure}[b]{0.239\textwidth}
         \centering
         \includegraphics[width=\textwidth]{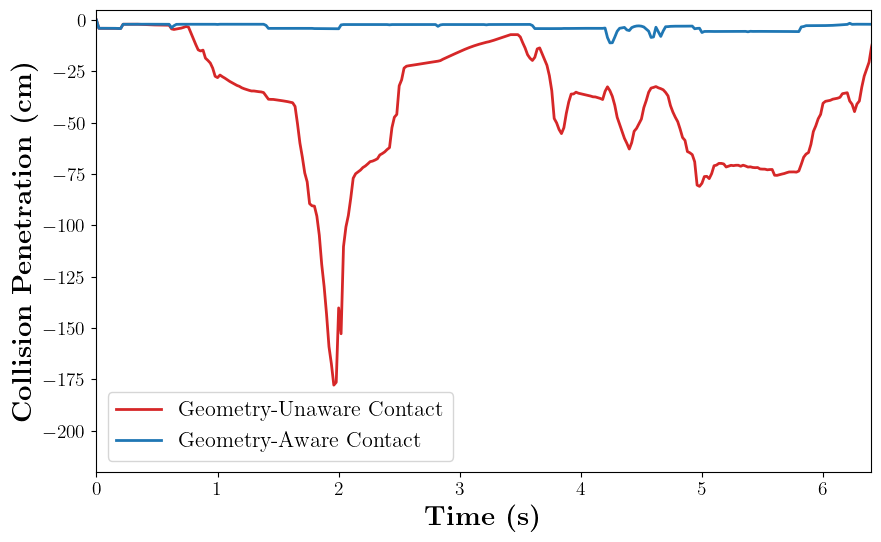}
         \caption{S1 w/o collision penalties}
     \end{subfigure}

     \vspace{3pt}
     
     \begin{subfigure}[b]{0.239\textwidth}
         \centering
         \includegraphics[width=\textwidth]{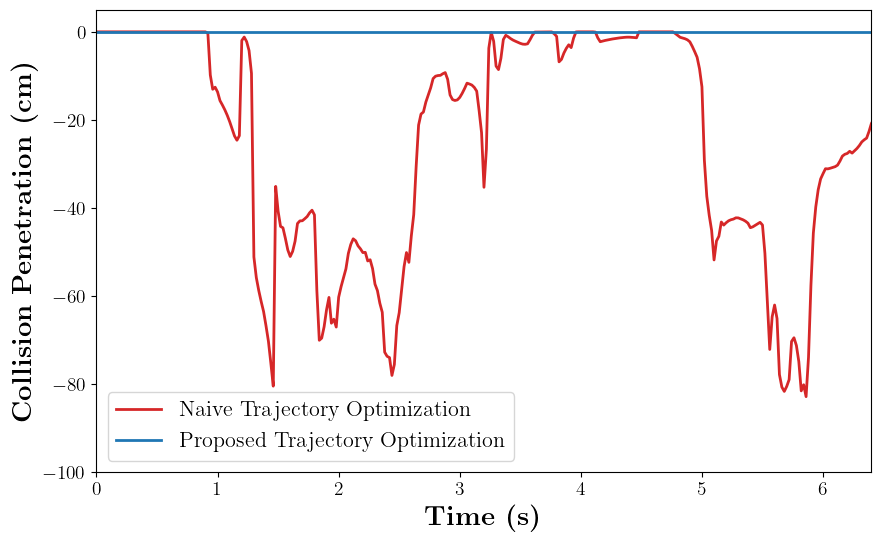}
         \caption{S2 w/ collision penalties}
     \end{subfigure}
     \begin{subfigure}[b]{0.239\textwidth}
         \centering
         \includegraphics[width=\textwidth]{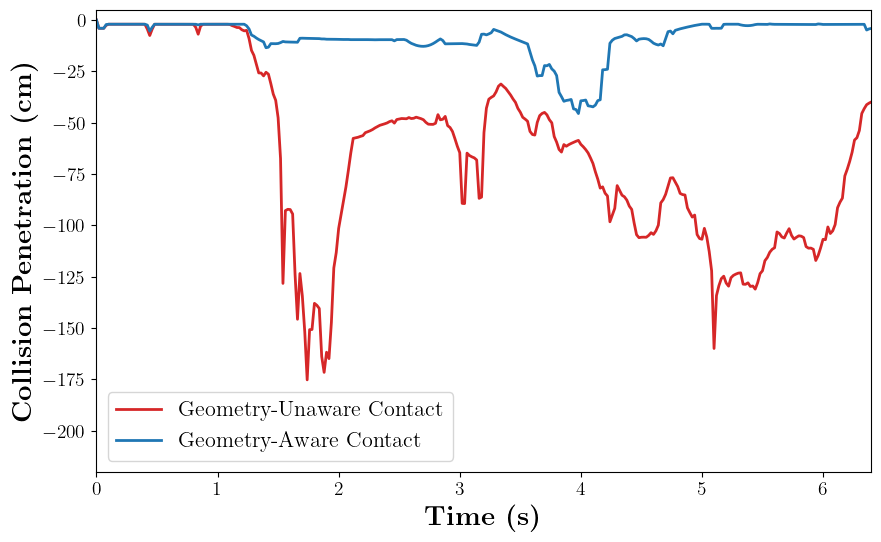}
         \caption{S2 w/ collision penalties}
     \end{subfigure}

    \caption{Collision penetrations comparison between the TO output using our proposed pipeline (blue) and a naive approach (red) for both the laundry scenario (S1) and the trolley scenario (S2) with and without collision penalties enabled.}
    \vspace{-7mm}
    \label{fig:collision_ablation}
\end{figure}

Figure~\ref{fig:collision_ablation} presents the total amount of negative collision penetrations at each timestep of the trajectory from MuJoCo, while enabling collision penalties for both scenarios. Our proposed pipeline maintains a collision-free behavior throughout the whole trajectory, while a naive approach experiences significant negative penetrations during the whole trajectory. One might argue that all possible collision constraints could be enabled in the TO to obtain a collision-free motion. However, in such a case TO fails to solve the task and gets stuck in a local minima.

\subsection{Geometry Improves Contact Transfer}
\label{sec:contact_ablation}

We present a comparison between the trajectory outputs when using our proposed contact transfer (Sec.~\ref{sec:contact_transfer}) and a geometry-unaware contact extraction process. Our contact extraction approach uses semantic and geometry cues from a semantic-aware foundation model and the objects' point clouds to refine incorrect semantic matches. On the other hand, the geometry-unaware contacts directly use the output of the model without any correction. Our metric for the comparison is the amount of collision penetration, both in terms of self-collisions and robot-object collisions. To study the effect of the contact extraction in isolation, we disable all collision penalties in the TO problem and use the same extracted robot configuration (Sec.~\ref{sec:retargeting}) in both approaches.

Figure~\ref{fig:collision_ablation} shows the result of the ablation study for scenarios S1 and S2. For both S1 and S2, we clearly see that a geometry-unaware contact transfer leads to a higher number of negative collisions, and therefore a higher amount of negative penetration during the trajectory. While with our approach, there still exists some negative penetrations but these are substantially less for both scenarios and can be prevented with a minimal set of collisions (Sec.~\ref{sec:physically_plausible_trajectories}). However, to obtain a collision-free motion, a geometric-unaware approach would require significantly more collision constraints which makes the problem extremely non-convex with many local minima. Local TO in such cases gets stuck in local minima and is unable to solve the task.

% \begin{figure}
%     \centering
%     \begin{subfigure}[t]{0.4\textwidth}
%         \includegraphics[width=\textwidth]{images/laundry_contact.png}
%         \caption{S1 collision data.}
%         \label{fig:contact_ablation_penetration_laundry}
%     \end{subfigure}
%     \begin{subfigure}[t]{0.4\textwidth}
%         \centering
%         \includegraphics[width=\textwidth]{images/trolley_contact.png}
%         \caption{S2 collision data}
%         \label{fig:contact_ablation_penetration_trolley}
%     \end{subfigure}
%     \caption{Comparison between the collision penetrations incurred by the TO outputs when using our proposed geometry-aware contact transfer (blue) and a geometry-unaware contact extraction (red) for both the laundry scenario (S1) and the trolley scenario (S2). In both cases, we use the same extracted robot configurations and we disable the collision penalties.}
%     \vspace{-7mm}
%     \label{fig:contact_ablation_penetration}
% \end{figure}

\subsection{Keyframes Reduce Penetration}
\label{sec:keyframe_ablation}

\begin{figure*}[t]
    \centering
    \begin{subfigure}[t]{\textwidth}
        \centering
        \includegraphics[width=.95\textwidth]{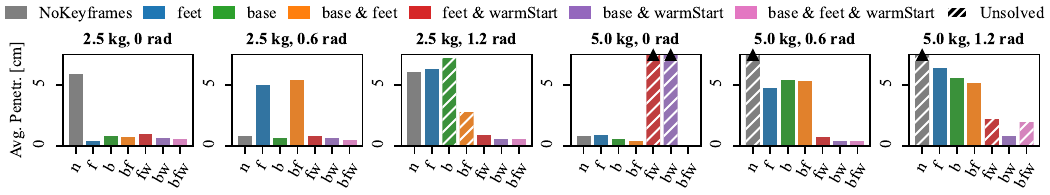}
        \caption{S1 collision data. The y-axis bound of average penetration is 7.5cm.}
        \label{fig:demo1_collision_kf_abl}
    \end{subfigure}
    \begin{subfigure}[t]{\textwidth}
        \centering
        \includegraphics[width=.95\textwidth]{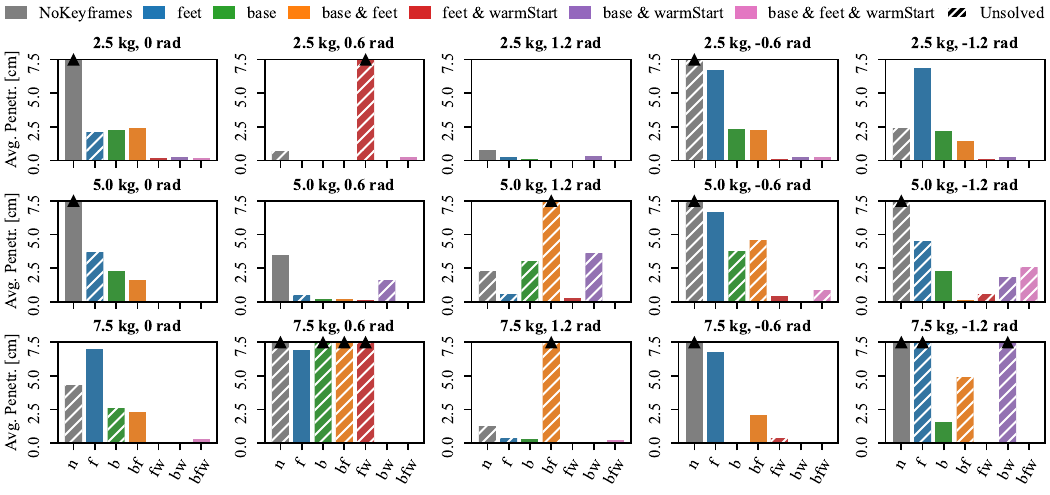}
        \caption{S2 collision data. The y-axis bound of average penetration is 15cm.}
        \label{fig:demo2_collision_kf_abl}
    \end{subfigure}
    \caption{Collision ablation study data. Subtitles are (box mass, robot initial yaw). The upward solid triangles denote that the average penetrations exceed the y-axis bound. Slash hatches on bars mean the optimization failed to converge with the corresponding settings. The legends represent settings of keyframe utilization. The average penetration is calculated as the sum of penetration divided by the number of shooting nodes. 
    % Note that in our TO, the robot hands are simplified as cylinders to reduce unnecessary computational complexities. Therefore, we do not compute the penetration of the simplified hands into the objects, as well as the foot bottom contact with the ground during stance and walking.
    }
    \vspace{-7mm}
    \label{fig:collision_kf_ablation}
\end{figure*}

In this section, we conduct ablation studies between the keyframe-guided TO and the TO without keyframes (we call this baseline NoKeyframe). Note that in this case, we use the refined contact information (Sec.~\ref{sec:contact_transfer}) in all problems and the focus is to quantify the effect of the different components of the keyframe. To do that, we formulate various settings with different combinations of keyframe utilization. This includes the keyframe base cost in~\eqref{eq:kf_base_costs}, the keyframe foot position cost as in~\eqref{eq:kf_foot_costs}, and the subtask-like warm start in Sec.~\ref{sec:kf_warm_start}. All formulations are provided with the same contact sequences and the same set of collision constraints as in Sec.~\ref{sec:physically_plausible_trajectories}.

\subsubsection{S1 (Basket retrieval)}
\label{sec:scenario_1}

Three keyframes are used in this experiment: picking up the box, placing the box, and grabbing the basket while standing on the box. We also varied in the experiment the box mass in the range $\{2.5, 5.0\}$kg, and the initial robot yaw angle $\phi$ in the range $\{0, 0.6, 1.2\}$rad. We have observed in our experiments that these two parameters have a large impact on the performance of TO. We added only a minimal set of collision constraints as defined in Sec.~\ref{sec:physically_plausible_trajectories} for S1. The results are shown in Fig.~\ref{fig:demo1_collision_kf_abl}.

\subsubsection{S2 (Trolley pushing)}
\label{sec:scenario_2}

In this scenario, to describe the goal of pushing the trolley forward, an additional cost on the trolley position is added. Three keyframes are used: picking up the box, placing the box, and the beginning of pushing the trolley. The set of tested box mass is $\{2.5, 5.0, 7.5\}$kg, and that of the initial robot yaw angle $\phi$ is $\{0, 0.6, 1.2, -0.6, -1.2\}$rad. A minimal set of collision constraints are are added as explained in Sec.~\ref{sec:physically_plausible_trajectories} for S2. The results are shown in Fig.~\ref{fig:demo2_collision_kf_abl}.

\subsubsection{Discussion}

For both S1 and S2, it can be observed in Fig.~\ref{fig:collision_kf_ablation} that using keyframes helps the success rate of TO in solving the problem, and leads to better solutions with low penetration. This is evident when we compare the case where AllKeyframe is used (pink, rightmost bars) versus the NoKeyframe case (grey, leftmost bars).

For example, in the (5.0kg, 0rad) test of S1, the AllKeyframe case results in zero penetration, while removing any part of the keyframe either hinders convergence or results in large penetrations. In particular, settings with the warm-start scheme from Sec.~\ref{sec:kf_warm_start} tend to have better convergence and lower penetration. In some cases, we can see the AllKeyframe case has achieved slightly worse results than other cases. We believe this is due to some details in the trajectory optimization solver.

\section{Conclusion} 
\label{sec:conclusion}

In this work, we presented a novel approach that generates physically consistent trajectories for long-horizon loco-manipulation tasks. We do so by leveraging LDMs that are able to synthesize high-quality images demonstrating how a human would accomplish a task. From such demonstrations, we extract the robot configurations and contact locations for a long-horizon high-level plan, which are used to guide a whole-body TO. We evaluated the proposed method in simulation for two challenging scenarios that require long-horizon reasoning, and showed that our proposed pipeline can generate physically plausible trajectories for long-horizon humanoid loco-manipulation tasks.

Future work will focus on implementing the proposed work on a real humanoid robot and evaluating its performance in the real world. Furthermore, we will relax the assumption of being given the high-level plan, thereby developing a long-horizon task planner that can reason based on visual feedback using recent advances in multimodal large language models. 
% Finally, we will improve the contact transfer limitation when dealing with extreme occlusions and lack of texture using a hypotheses system based on the semantic scores.

%                                   % the textheight of the last page by a suitable amount.
%                                   % This command does not take effect until the next page
%                                   % so it should come on the page before the last. Make
%                                   % sure that you do not shorten the textheight too much.

%%%%%%%%%%%%%%%%%%%%%%%%%%%%%%%%%%%%%%%%%%%%%%%%%%%%%%%%%%%%%%%%%%%%%%%%%%%%%%%%

%%%%%%%%%%%%%%%%%%%%%%%%%%%%%%%%%%%%%%%%%%%%%%%%%%%%%%%%%%%%%%%%%%%%%%%%%%%%%%%%

%%%%%%%%%%%%%%%%%%%%%%%%%%%%%%%%%%%%%%%%%%%%%%%%%%%%%%%%%%%%%%%%%%%%%%%%%%%%%%%%
\bibliographystyle{IEEEtran}
\bibliography{bibliography, opt}

\end{document}